\documentclass[letterpaper, 10 pt, journal, twoside]{IEEEtran}

\IEEEoverridecommandlockouts            

\usepackage{epsfig} 
\usepackage{mathptmx} 
\usepackage{times} 
\usepackage{amsmath} 
\usepackage{amssymb}  
\usepackage{bm}
\usepackage{soul}

\usepackage{cite}
\usepackage{color}

\usepackage[ruled,vlined]{algorithm2e}
\usepackage{array}

\usepackage{graphicx}
\usepackage{subcaption}
\usepackage{mwe}
\usepackage{comment}

\usepackage[labelsep=endash]{caption}

\title{\huge
Online Target Localization using Adaptive Belief Propagation in the HMM Framework
}

\author{Min-Won Seo$^{1}$ and  Solmaz S. Kia$^{1}$, \emph{Senior Member, IEEE}
\thanks{Manuscript received: February, 24, 2022; Revised: April, 24, 2022; Accepted: June, 29, 2022.}%
\thanks{This paper was recommended for publication by
Editor Sven Behnke upon evaluation of the Associate Editor and Reviewers’
comments. This work was supported by NIST award 70NANB17H192.}
\thanks{$^{1}$Min-Won Seo, and Solmaz S. Kia are with the Department of Mechanical and Aerospace Engineering, University of California, Irvine, CA 92697, USA,
{\tt\footnotesize\{minwons,kia\}@uci.edu}}%
\thanks{Digital Object Identifier (DOI): see top of this page.}
}

\newcommand{\real}{{\mathbb{R}}}

\begin{document}

\markboth{}
{SEO AND KIA: Online Target Localization using Adaptive Belief Propagation in the HMM Framework}

\maketitle

\begin{abstract}
This paper proposes a novel adaptive sample space-based Viterbi algorithm for target localization in an online manner. The method relies on discretizing the target's motion space into cells representing a finite number of hidden states. Then, the most probable trajectory of the tracked target is computed via dynamic programming in a Hidden Markov Model (HMM) framework. The proposed method uses a Bayesian estimation framework which is neither limited to Gaussian noise models nor requires a linearized target motion model or sensor measurement models. However,  an HMM-based approach to localization can suffer from poor computational complexity in scenarios where the number of hidden states increases due to high-resolution modeling or target localization in a large space. To improve this poor computational complexity, this paper proposes a belief propagation in the most probable belief space with a low to high-resolution sequentially, reducing the required resources significantly. The proposed method is inspired by the $k$-d Tree algorithm (e.g., quadtree) commonly used in the computer vision field. Experimental tests using an ultra-wideband (UWB) sensor~network demonstrate our results.
\end{abstract}


\IEEEpeerreviewmaketitle

\section{Introduction}
\label{sec::Introduction}

\begin{figure}[!t]
    \centering
    \includegraphics[width=0.48\textwidth]{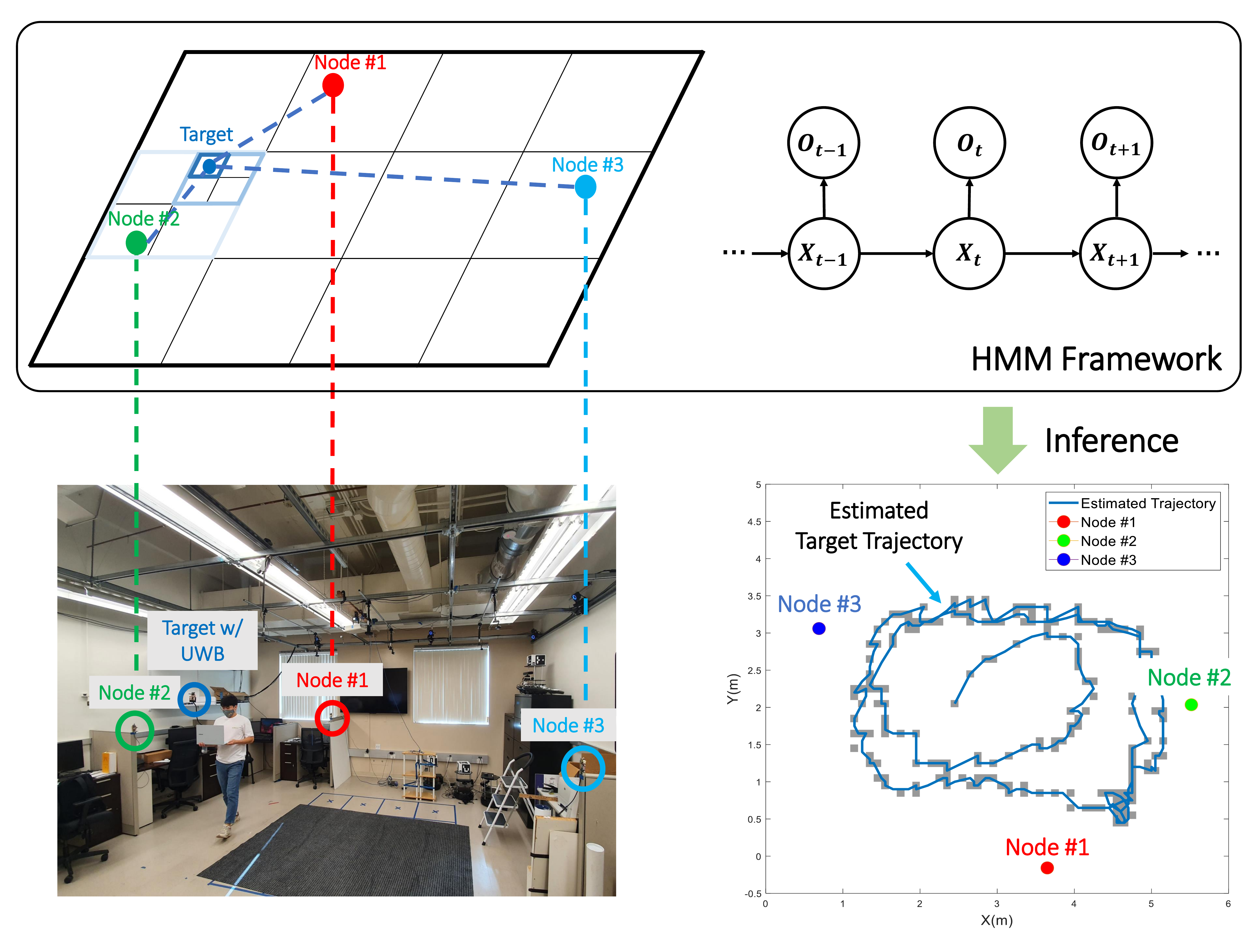}
    \caption{{\small UWB-based target localization problem definition: the trajectory of an uncertain mobile agent, equipped with a UWB tag (e.g., smartphone), is estimated in UWB sensor networks.}}
    \label{fig:Prob_Def}
\end{figure}

\IEEEPARstart{L}{ocalization} has a pivotal role in the successful execution of fundamental tasks such as planning and control of autonomous mobile agents. Despite advances in various localization algorithms and sensor technologies, indoor localization still remains a challenging task. Indoors, Global Positioning System (GPS) due to weak signal strength fails to provide reliable localization. 
In recent years, wireless signal-assisted  localization techniques, e.g., using Wi-Fi signal~\cite{biswas2010wifi, Yoo2016Wifi}, Bluetooth~\cite{raghavan2010accurate, han2018hmm}, and ultra-wideband (UWB)  \cite{nguyen2018UWB,cao2020accurate,Zhu2021UWB}, have emerged as effective indoor localization solutions. These techniques typically use pre-installed devices (beacons) with known locations. They rely on  time-of-arrival (TOA) and received signal strength (RSS) measurements to obtain a relative distance of the mobile agent from the pre-installed beacons and use this relative range measurement and the known location of the beacons to localize the mobile agent. Among these wireless technologies, UWB, due to its high time resolution, large signal bandwidth, and capability to operate under none-line-of-sight (NLoS) conditions, has received a lot of attention. UWB transceivers are now being also embedded on smartphones~\cite{2019Iphone} and are readily  available for localization purposes. In general, however, the presence of unwanted reflections and obstacles causes signal perturbation and errors in estimation results. In this paper, we consider the problem of UWB-based mobile agent localization, and to improve the resolution and accuracy of the localization we propose a Hidden Markov Model (HMM) based framework to track moving targets by decoding motion~trajectory, see Fig.~\ref{fig:Prob_Def}. The main contribution of this paper is addressing the memory and computation cost of an HMM-based localization via adaptive belief propagation.

Recent literature has shown HMM frameworks can provide effective and promising solutions to localization, tracking, and search problems \cite{morelli2007hidden, seitz2010hidden, liu2012hybrid, Nisar2015Uav, bayoumi2019speeding, rudic2020geometry, vandermeulen2020sampling}. In the HMM-based approaches, the discretized position (i.e., discrete state-space) is defined as a hidden state and is estimated using observation data (i.e., sensor measurements). Here, probabilities can be expressed explicitly based on discrete sample space, thus performing Bayesian inference efficiently without Gaussian assumption and linearization. The author in \cite{morelli2007hidden} proposes a grid-based Bayesian method to jointly track the sequence of the positions of the target and channel condition identification in UWB sensor networks. In \cite{seitz2010hidden, liu2012hybrid}, the HMM is used to tackle data fusion of inertial sensor information and Wi-Fi signals which cannot be handled properly under Gaussian assumption. Furthermore, when system and/or sensor models cannot be defined properly, in other words, for highly uncertain systems, the HMM can offer a pragmatic solution. HMM has also been used in~\cite{Nisar2015Uav} to deal with highly uncertain target motions and fusion with uncertain sensor models on the road network-based target tracking. The effectiveness of HMM framework to incorporate geometric information such as barriers or boundaries into the localization problem has been demonstrated in~\cite{rudic2020geometry}.

Despite the great promise of discrete-state HMM-based approaches for mobile agent localization, the high computational complexity stays a major challenge in the way of adoption of such approaches for real-time localization. To be precise, the computational complexity is  quadratic in the number of hidden states, which limits the application to problems with fine-grid discretization or high-performance computing systems. Moreover, high memory size is required to implement the Viterbi algorithm, a main component of the HMM approaches. The Viterbi algorithm needs to store the sequence of positions at every time step. Therefore, the number of hidden states (sample space) is important to implement its algorithm in an efficient and online manner. To tackle these difficulties, \cite{trogh2015advanced, scherhaufl2018blind, sun2019practical, bayoumi2019speeding} represented their environments with a few discretized areas. The
author in \cite{sun2019practical} classified the floor area into three categories such as rooms, corridors, and entrance/exits. \cite{bayoumi2019speeding} used a grid map representation with a few overlaid graphs. In \cite{champlin2000target, chigansky2011viterbi, rudic2020discrete}, continuous-state-based Viterbi algorithm as an alternative approach is proposed instead of discrete-state. However, computational issues still remain in \cite{champlin2000target, chigansky2011viterbi}. Although \cite{rudic2020discrete} proposed a sampling-based method to reduce computation complexity which is linear in the number of samples, there are Gaussian assumptions for sampling and decoding the sequence of positions.

In this paper, we present an efficient method for reducing the computational cost of the discrete state HMM-based localization. Inspired by the $k$-d Tree algorithm (e.g., quadtree)\cite{vedaldi2010vlfeat} that is commonly used in the computer vision field, we propose a belief propagation in the most probable belief space with a low to high-resolution sequentially, thus reducing the required resources significantly. Our method has three advantages for localization: (a) since finite grid sample space represents the discrete distribution,  there are no Gaussian assumptions and linearization; (b) it can handle the whole area of interest, not specific or small map representations; (c) The required resources to enable an HMM modeling are reduced significantly via our proposed adaptive scheme. We demonstrate our results via a set of experiments to localize the trajectory of a human agent which is equipped with only a UWB transceiver.

The remainder of this paper is organized as follows: Section~\ref{sec::Target_Loc} presents the problem definition. Section~\ref{sec::Adaptive_Viterbi} describes the HMM framework, HMM-based trajectory estimation, and adaptive sample space-based Viterbi algorithm. Lastly, a UWB-based localization application is described. Section~\ref{sec::ExpResult} reports the experimental results. Finally, in Section~\ref{sec::Con}, the conclusions and future works are drawn.

\section{Problem Definition}
\label{sec::Target_Loc}
Consider a localization problem in which a mobile agent (hereafter referred to as target) equipped with a UWB transceiver (e.g., a pedestrian with a smartphone) moves in an environment with 
a UWB sensor network with $K\geq 3$ anchors located at $P_k\in\real^2$, $k\in\{1,\dots,K\}$, as depicted in Fig.~\ref{fig:Prob_Def}.
The anchors are either previously installed nodes or mobile nodes which transmit their position information. Let us denote the position of the target at time $t$ as $\mathbf{p}_t=[x,\;y]^\top$, where $x$ and $y$ represent its 2D coordinates at time index $t$. When the target enters the coverage area of the UWB sensor networks, its UWB sensor can communicate with the anchor nodes and obtain relative range measurements. Our goal is to estimate the most probable sequence of the position of the target, $\hat{\mathbf{p}}_{0:t}$, using the UWB measurements without prior knowledge of the locomotion model, its initial position, and without the use of any proprioceptive sensors such as IMUs or encoders. To meet our goal we rely on a discrete-time discrete-state (DTDS) HMM framework.

\begin{figure}[!t]
    \centering
    \includegraphics[width=0.47\textwidth]{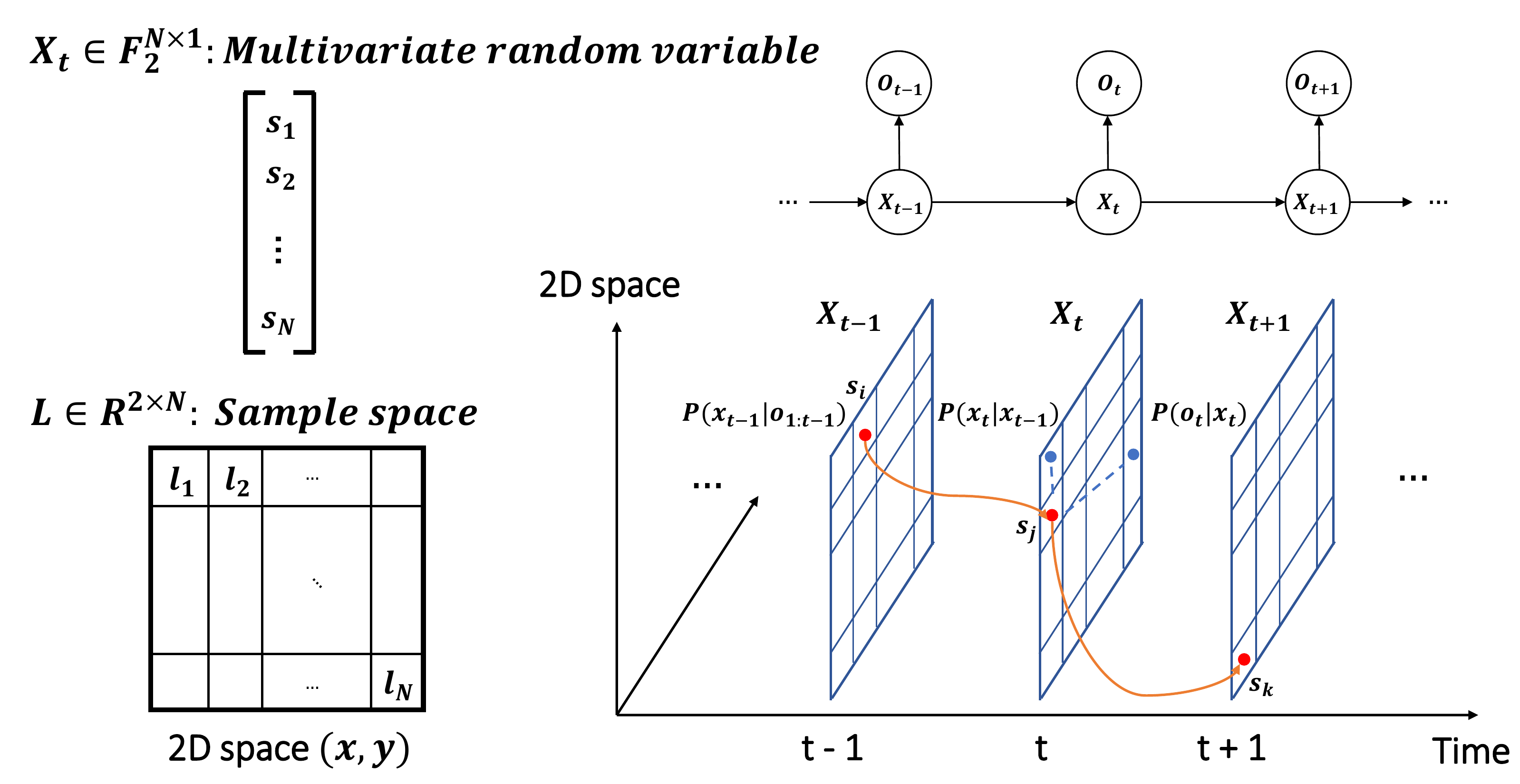}
    \caption{{\small Localization via Hidden Markov Model. $\mathbf{X}_t$ represents multivariate random variable at every time step. Each element $s_i$ of $\mathbf{X}_t$ is mapped into the cell $l_i$ 
    in 2D sample space $\mathbf{L} \in \mathbb{R}^{2 \times N}$ at time index $t$. The recursive Bayesian estimation is implemented via transition probability $\mathbb{P}(\mathbf{x}_t|\mathbf{x}_{t-1})$ and observation probability $\mathbb{P}(\mathbf{o}_t|\mathbf{x}_t)$.}}
    \label{fig:HMM_Loc}
\end{figure}

\section{Adaptive Sample Space-based Belief Propagation for Trajectory estimation}
\label{sec::Adaptive_Viterbi}
\subsection{Hidden Markov Model Framework and Parameters}
\label{sec::Loc_HMM}

To develop our DTDS HMM-based localization algorithm, we uniformly discretize the work-space of the target to generate a 2D sample space $\mathbf{L} \in \mathbb{R}^{2 \times N}$, where $N$ is the total number of discretized cells. A finite number of multivariate random variables represent the uniformly discretized areas of interest, and the Hidden Markov process is implemented as shown in Fig.~\ref{fig:HMM_Loc}. 
The DTDS HMM has two stochastic processes, hidden states, $\mathbf{x}_t \in \mathbf{X}_t$, and observations, $\mathbf{o}_t \in \mathbf{O}_t$ at each time step $t$.
The hidden state represents the belief of the target's position in discretized sample space. Observations are obtained from UWB anchors. While observations $\mathbf{o}_t$ can take real values, the hidden state $\mathbf{x}_t$ takes values from $\mathbf{X}_t = [s_1,s_2,\dots,s_N]^\top$,  $\mathbf{X}_t \in \mathbb{F}_2^{N \times 1}$, where $s_i$ takes the binary number $1$ (the target exists in the cell) or $0$ (the target does not exist). Each element $s_i$ is mapped into the grid cell $l_i$ in 2D sample space $\mathbf{L} \in \mathbb{R}^{2 \times N}$. Since we are localizing only one target, at each time step $t$, $\mathbf{X}_t$ can have \emph{only one} $s_i=1$, and \emph{all the other states are set to} $s_j\!=\!0, \; j \neq i$. In what follows all multivariate random variables are written in uppercase bold letters using a time index $t$. $\mathbb{P}(\cdot)$ denotes the probability mass function (PMF) for discrete state space 
or the probability density function (PDF) for continuous states, e.g., $\mathbb{P}(\mathbf{X}_t)$ is the PMF of  $\mathbf{X}_t$, while $\mathbb{P}(\mathbf{x}_t)$ is the probability of $\mathbf{X}_t = \mathbf{x}_t$. The same notation is used for the conditional mass function, e.g., $\mathbb{P}(\mathbf{X}_t| \cdot )$. In addition, $\hat{x}_t$ refers to an estimated state, and $x_t^g$ is the high precision position in 2D space. Note that $\mathbf{L}\hat{x}_t$ corresponds to the target location estimate $\hat{\mathbf{p}}_t$.  

In the HMM-based localization, there are two fundamental elements that should be defined: the \emph{transition probability}, $\mathbb{P}(\mathbf{x}_t|\mathbf{x}_{t-1})$, which indicates the probability of transitioning from hidden state $\mathbf{x}_{t-1}$ to $\mathbf{x}_t$, and the \emph{observation probability}, $\mathbb{P}(\mathbf{o}_t|\mathbf{x}_t)$, which indicates the probability of measurement $\mathbf{o}_t$ given hidden state $\mathbf{x}_t$. Given these probability models, the recursive Bayesian estimation procedure processes the observation data obtained from the sensor network to produce an estimate on the trajectory of the target. The HMM-based localization algorithm we introduce in Section~\ref{sec::Adaptive_Viterbi} is a generalized framework that makes no assumption on   $\mathbb{P}(\mathbf{x}_t|\mathbf{x}_{t-1})$ and $\mathbb{P}(\mathbf{o}_t|\mathbf{x}_t)$.

\begin{figure}[!t]
    \centering
    \includegraphics[width=0.47\textwidth]{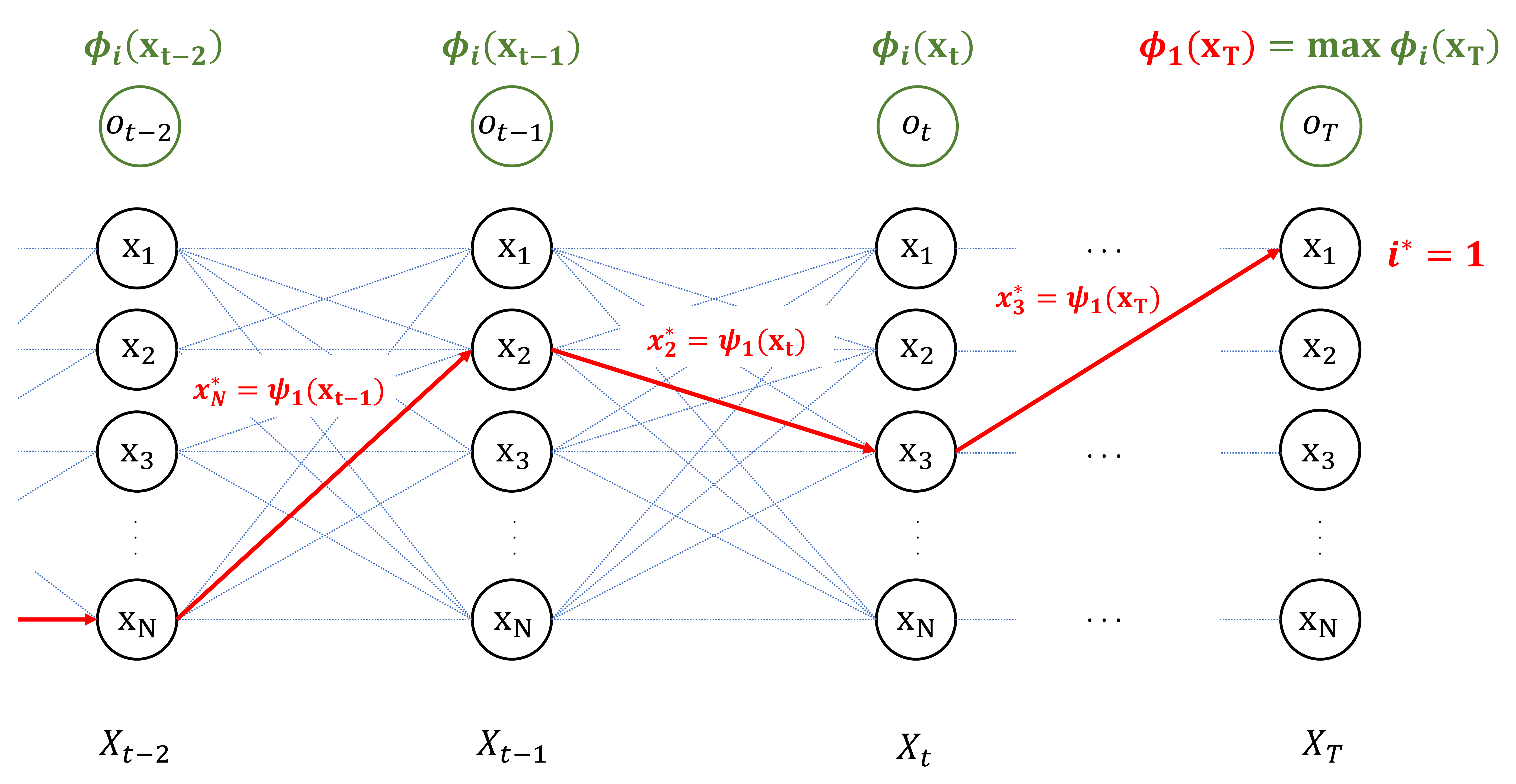}
    \caption{{\small The schematic representation of the Viterbi algorithm. $\psi_i(\mathbf{x}_t)$(or $\Psi_i(\mathbf{x}_t)$) and $\phi_i(\mathbf{x}_t)$ (or $\rho_i(\mathbf{x}_t)$) are stored at every step. In this example, since maximum value of $\phi_i(\mathbf{x}_T)$ is $\phi_{i^{\ast} = 1}(\mathbf{x}_T)$ at time $T$, the MAP sequence $\hat{\mathbf{x}}_{0:T}$ is decoded from the memorization table $\Psi_{i^\ast = 1}(\mathbf{x}_{0:T})$.}}
    \label{fig:Viterbi_dig}
\end{figure}

\subsection{Hidden Markov Model-Based Trajectory Estimation}
HMM-based estimators can be designed to estimate either the joint posterior, $\mathbb{P}(\mathbf{X}_{0:t}|\mathbf{O}_{1:t})$, or the marginal posterior, $\mathbb{P}(\mathbf{X}_{t}|\mathbf{O}_{1:t})$. Here we develop an HMM-based localization method that estimates 
That is, we find the most likely trajectory of the target from the initial time $0$ to the current time $t$, $\mathbb{P}(\mathbf{x}_{0:t}|\mathbf{o}_{1:t})$, rather than the most probable position at a certain time step $t$, $\mathbb{P}(\mathbf{x}_{t}|\mathbf{o}_{1:t})$. In other words, the goal is to find a MAP sequence (trajectory) of the target. Hence, we do not need to calculate the sum and/or integral for marginalization or to compute the actual belief values of the distribution. It means that we can use relative values of the beliefs and thus omit normalizing factors in every probability multiplication calculation. In addition, MAP delivers a better localization performance than the minimum mean square error (MMSE) for the target tracking problems, which may have multimodal posterior~\cite{bar1995multitarget, blom2008tracking}. The well-known Viterbi algorithm offers a computationally efficient procedure for finding this MAP sequence via dynamic programming~\cite{svensen2007pattern}, see Fig.~\ref{fig:Viterbi_dig}.

To obtain the most likely trajectory of the target from the initial time $0$ to the current time $t$, $\mathbb{P}(\mathbf{x}_{0:t}|\mathbf{o}_{1:t})$ we apply the Viterbi algorithm~\cite{svensen2007pattern}, as explained below. At each time step $t > 1$, for each given $\mathbf{x}_i$, $i\in\{1,\cdots,N\}$, the most probable state sequence $\psi_i(\mathbf{x}_t)$ is obtained from
\begin{align}
\label{eq::Viterbi_eq1}
    \mathbf{x}_{t-1}^{\ast} = 
    \arg \! \max_{\mathbf{x}_{t-1}} \; \mathbb{P}(&\mathbf{X}_t = \mathbf{x}_{i}\,|\, \mathbf{X}_{t-1} = \mathbf{x}_{t-1}) \mathbb{P}(\mathbf{x}_{0:t-1}\,|\,\mathbf{o}_{1:t-1}), \\
\label{eq::Viterbi_eq1_1}
    &\psi_i(\mathbf{x}_t) \triangleq \:  \arg \! \max_{s_j} \mathbf{x}_{t-1}^{\ast}
\end{align}
where $\mathbb{P}(\mathbf{X}_t=\mathbf{x}_i\,|\, \mathbf{X}_{t-1}=\mathbf{x}_{t-1}) =  \mathbb{P}(\mathbf{X}_t = \mathbf{x}_{i}\,|\, \mathbf{X}_{0:t-1} = \mathbf{x}_{0:t-1})$ follows the Markov assumption. It is known that the computational complexity of solving~\eqref{eq::Viterbi_eq1} is $O(N^2)$. $\psi_i(\mathbf{x}_t)$ takes an element $s_j=1$ from $\mathbf{x}_{t-1}^{\ast}$, and the joint posterior $\phi_i(\mathbf{x}_{t})$ is obtained from
\begin{align}
\label{eq::Viterbi_eq2}
    \phi_i(\mathbf{x}_t) \triangleq 
    \: \mathbb{P}(\mathbf{o}_t \,&|\, \mathbf{X}_t=\mathbf{x}_i) \mathbb{P}( \mathbf{X}_t=\mathbf{x}_i \, |\, \mathbf{x}_{t-1}^{\ast}) \mathbb{P}(\mathbf{x}_{0:t-1}^{\ast}\,|\,\mathbf{o}_{1:t-1}), \\
\label{eq::Viterbi_eq3} 
    &\phi_i(\mathbf{x}_t) \propto \mathbb{P}(\mathbf{x}_i,\mathbf{x}_{0:t-1}^{\ast}\,|\,\mathbf{o}_{1:t}),
\end{align}
which are evaluated and then stored. These processes are implemented recursively until final time $T$. At the first step $(t = 1)$, the most probable state and a joint posterior are directly obtained from the maximum likelihood as \eqref{eq::Viterbi_eq4} and \eqref{eq::Viterbi_eq5}, respectively,
\begin{align}
\label{eq::Viterbi_eq4}
    \psi_i(\mathbf{x}_t) =  \arg \! \max_{s_j} \{ \arg \! \max_{\mathbf{x}_t}\mathbb{P}(\mathbf{o}_t\,|\,\mathbf{X}_t = \mathbf{x}_t) \},  \quad \forall i, 
\end{align}

\begin{align}
\label{eq::Viterbi_eq5}
    \phi_i(\mathbf{x}_t) = \mathbb{P}(\mathbf{o}_t \, | \, \mathbf{x}_t^*), \quad \forall i.
\end{align}
In order to avoid numerical issues (underflow) resulting from multiplying probabilities recursively in the implementation, the logarithm is applied to the calculation, i.e. $\rho_i(\mathbf{x}_t) := \ln \phi_i(\mathbf{x}_t)$. The $\ln()$ function is strictly monotonic, and $\ln(\prod f(x)) = \sum \ln f(x)$ holds. Therefore, equations \eqref{eq::Viterbi_eq1_1} and \eqref{eq::Viterbi_eq3} convert into \eqref{eq::Viterbi_eq7} and \eqref{eq::Viterbi_eq8}, respectively,
\begin{align}
\label{eq::Viterbi_eq7}
   \Psi_i(\mathbf{x}_t) \: \triangleq 
    \: \arg \! \max_{s_j}\{\arg \! \max_{\mathbf{x}_{t-1}} \; \ln \mathbb{P}(\mathbf{X}_t & = \mathbf{x}_i\,|\, \mathbf{X}_{t-1} = \mathbf{x}_{t-1}) \\ \nonumber
    \;+\; &\ln \mathbb{P}(\mathbf{x}_{0:t-1}\,|\,\mathbf{o}_{1:t-1})\},
\end{align}
\begin{align}
\label{eq::Viterbi_eq8}    
    \rho_i(\mathbf{x}_t) \propto \ln \mathbb{P}(\mathbf{x}_i,\mathbf{x}_{0:t-1}^{\ast}\,|\,\mathbf{o}_{1:t}).
\end{align}
Finally, the MAP sequence $\hat{\mathbf{x}}_{0:T}$ is decoded from the memorization table, $\Psi_{i^\ast}(\mathbf{x}_{0:T})$, by backtracking process from $i^{\ast} \in \{1, \dots, N\}$ as follows
\begin{align}
\label{eq::Viterbi_eq6}
   s_{i^{\ast}}  = \arg \! \max_{s_i} \{\arg \! \max_{\mathbf{x}_T} \rho_i(\mathbf{x}_T) \}.
    \end{align}

\begin{figure}[!t]
    \centering
    \includegraphics[width=0.47\textwidth]{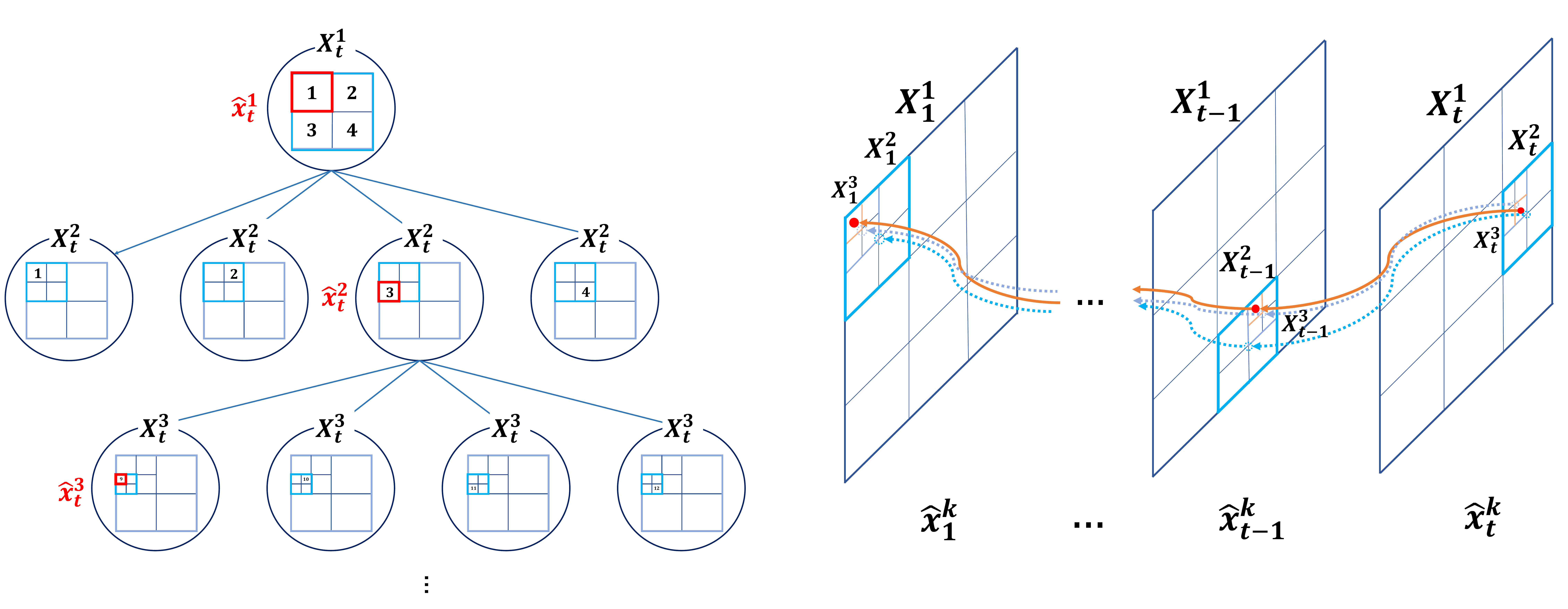}
    \caption{{\small Adaptive sample space-based sequential Viterbi algorithm. Based on the quadtree algorithm, the belief propagation area of the agent is changed from low $\mathbf{X}_t^1$ to high $\mathbf{X}_t^r$ resolution space. The position of the agent $\hat{\mathbf{x}}_{t}^k$ is estimated according to the sample space $\mathbf{X}_t^k$ sequentially.}}
   \label{fig:Sqt_Viterbi}
\end{figure}

\medskip
\subsection{Adaptive Sample Space-based Belief Propagation}
\label{sec::QuadTree}
The size of the hidden state space, related to resolution, has an effect on estimation accuracy and computation time. Obviously, there is a trade-off between these two factors, i.e., high resolution increases estimation accuracy, but computation efficiency is reduced drastically, and vice versa. In order to implement the Viterbi algorithm in an online manner, the number of hidden states is crucial. The key point of the sequential Viterbi algorithm is that we choose the most probable state based on relative probability for the whole hidden state, and thus we can adjust the hidden state space. We bring the concept of the $k$-d tree, commonly used in the computer vision field\cite{vedaldi2010vlfeat}, to adjust the size of the hidden state $\mathbf{X}_t$. As a result, the computation time of the Viterbi algorithm is decreased significantly while maintaining estimation accuracy. 

We use the quadtree, one of the $k$-d trees, to generate an adaptive sample space scheme in the 2D space in which we are solving our localization problem. Quadtree can be employed to represent a region of planar space. Quadtree decomposes the area of interest by recursively dividing it into four identical quadrants. In the case of a 3D localization problem, we can use octree in the same manner. Based on the property, the hidden state space is changed from low-resolution $u_1$ to high-resolution $u_r$. We define $r$-ordered resolution tuple $(u_1, \dots, u_r)$ as follows
\begin{align}
\label{eq::Viterbi_eq9}
    u_{k+1} = \frac{1}{2} \, u_k, \quad k \in \{1, \dots, r-1\}. 
\end{align}

In an adaptive sample space scheme, the hidden state $\mathbf{X}_t^k = [s_{1,t}^k,\, s_{2,t}^k,\, \dots,\, s_{N_k,t}^k]$, $\mathbf{X}_t^k \in \mathbb{F}_2^{N_k \times 1}$ is defined according to the resolution value $u_k$. Here, $N_k$ is the total element number of the hidden states $\mathbf{X}_t^k$. Once again since we are localizing only a single mobile agent, $\mathbf{X}_t^k$ can have only one $s_{i,\:t}^k=1$ and all other states are set to $s_{j,\:t}^k\!=\!0, \; j \neq i$. Therefore, we can express the relationship between $\mathbf{X}_t^k$ and $\mathbf{X}_t^{k-1}$.

At the first step, belief propagation in hidden state space $\mathbf{X}_t^1$ with resolution value $u_1$ is implemented according to equations \eqref{eq::Viterbi_eq7} and \eqref{eq::Viterbi_eq8}, and thus MAP sequence $\hat{\mathbf{x}}_{1:t}^1$ is defined between $\mathbf{x}_1^1$ and $\mathbf{x}_t^1$ from equation \eqref{eq::Viterbi_eq6}. Next step $(k \geqq 2)$, MAP sequence $\hat{\mathbf{x}}_{1:t}^k$ is found based on the belief area of previous estimated MAP sequence $\hat{\mathbf{x}}_{1:t}^{k-1}$ as depicted in Fig.~\ref{fig:Sqt_Viterbi}. Here, the relationship of elements between $\hat{\mathbf{x}}_{1:t}^{k}$ and $\hat{\mathbf{x}}_{1:t}^{k-1}$ is defined as
\begin{align}
\label{eq::Viterbi_eq10}
    s^{k-1}_{n,\;t} = \sum\nolimits_{l=4n-3}^{4n} s^k_{l,\;t}, \quad n \in \{1,\dots,N_{k-1}\}.
\end{align}
Then, we define a vector space set $\mathcal{Z}_t^k=[s_{1,t}^k,\dots,s_{N_k,t}^k]$ such that it only takes an element $s^k_{l,\;t}=1$. Here, $N_k\:(k \geqq 2)$ is four because of quadtree. As a result, at each step ($u_k \geqq u_2$), equations \eqref{eq::Viterbi_eq4}, \eqref{eq::Viterbi_eq7}, and \eqref{eq::Viterbi_eq8} are converted into \eqref{eq::Viterbi_eq11}, \eqref{eq::Viterbi_eq12}, \eqref{eq::Viterbi_eq13}, respectively as follows,

\begin{align}
\label{eq::Viterbi_eq11}
    \psi_i(\mathbf{x}_t^k) \: = \: \arg \! \max_{s_j^k} \{\arg \! \! \! \max_{\mathbf{x}_t^k \in \mathcal{Z}^k_t }\mathbb{P}(\mathbf{o}_{t}\,|\,\mathbf{X}_t^k = \mathbf{x}_t^k)\},
\end{align}

\begin{align}
\label{eq::Viterbi_eq12}
    \Psi_i(\mathbf{x}_t^k) \: =
    \: \arg \! \max_{s_j^k}\{\arg \!\!\!\!\!\!\!\! \max_{\mathbf{x}_{t-1} \in \mathcal{Z}^k_{t-1}} \ln \mathbb{P}(\mathbf{X}_t^k = \mathbf{x}_i^k\,|\, \mathbf{X}_{t-1}^k = \mathbf{x}_{t-1}^k)\nonumber \\ 
  \!\!\!\!\!\!\!\!  \;+\; \ln \mathbb{P}(\mathbf{x}_{0:t-1}^k\,|\,\mathbf{o}_{1:t-1})\}, \quad \forall \mathbf{x}_i^k \in \mathcal{Z}^k_t,
\end{align}

\begin{align}
\label{eq::Viterbi_eq13}    
    \rho_i(\mathbf{x}_t^k) \propto \ln \mathbb{P}(\mathbf{x}_i^k,\mathbf{x}_{0:t-1}^{k,\ast}\,|\,\mathbf{o}_{1:t}).
\end{align}

Through such procedure, the final estimation result is found as $\hat{\mathbf{x}}_{0:T}^{r}$, and thus the computation complexity for every time step is decreased from $O(N^2 \times T)$ to $O((N_1^2 + 4^2 \times (r-1)) \times T)$. Also, the space complexity is reduced from $O(N)$ to $O(N_1 + 4 \times (r-1))$. Here, $N \gg N_1$ generally. These complexities can have more advantages as much as finer discretization size $N$ and longer total time $T$.

\subsection{Application to UWB-based localization}
To localize a target that is not equipped with any proprioceptive sensor, such as IMUs or encoders, one can use the transition probability of 
\begin{align}
\label{eq::Trans_Prob}
    \mathbb{P}(\mathbf{x}_t|\mathbf{x}_{t-1} ; T_s, v_c) \propto \exp(-\frac{1}{2}\frac{(\|\mathbf{L}\mathbf{x}_t-\mathbf{L}\mathbf{x}_{t-1} \|_{2} - T_s v_c)^2} {\sigma_{x}^2}),
\end{align}
where $\sigma_{x}\in\real_{>0}$ denotes the standard deviation in the assumed motion model, $T_s$ is the inter-sampling time and $v_c\in\real_{\geq 0}$ is the constant velocity of the target. Here, $T_s$ and $v_c$ are predefined hyper-parameters of the model. When the velocity of the target is not known, one can set $v_c$ to zero in~\eqref{eq::Trans_Prob} and compensate for this lack of knowledge using a larger $\sigma_{x}$. The transition probability~\eqref{eq::Trans_Prob}, without the necessity for a proprioceptive sensor, gives the conditional probability of the position variable $\mathbf{L}\mathbf{x}_t$ given the previous position $\mathbf{L}\mathbf{x}_{t-1}$ by taking into account the natural temporal dependencies among the position variables due to the constant velocity.
 
In this paper, we consider line-of-sight (LoS) UWB range measurements and model the UWB range observation probability $\mathbb{P}(\mathbf{o}_t|\mathbf{x}_t)$ as 
\begin{align}
\label{eq::Obs_Prob1} 
\mathbb{P}(o_{t,k}|\mathbf{x}_t) \propto \exp(-\frac{1}{2}\frac{\|o_{t,k}\,-\,\theta_k(\mathbf{L}\mathbf{x}_{t})\|_{2}^2 }{\sigma_{o}^2}),
\end{align}
where $o_{t,k}$ is the measurement value from the $k$-th anchor at the time index $t$. $\theta_{k}(\mathbf{L}\mathbf{x}_t)$ is a function of Euclidean distances, i.e. $\theta_{k}(\mathbf{L}\mathbf{x}_t) = \Vert \mathbf{L}\mathbf{x}_{t} - P_{k} \Vert_{2}$, which makes the observation probability a nonlinear function. Our HMM-based localization algorithm handles this nonlinear measurement model without linearization. Here, $\sigma_{o}\in\real_{>0}$ is the standard deviation of the measurement $o_{t,k}$. Notice that for a given hidden state $\mathbf{x}_t$, the measurement model~\eqref{eq::Obs_Prob1} gives higher probability to a measurement when the measured distance $o_{t,k}$ is closer to the distance $\theta_{k}(\mathbf{L}\mathbf{x}_t)$ corresponding to state~$\mathbf{x}_t$.

\medskip

Let $K_{c,t}$ be the number of UWB anchors the target can connect to at time step $t$. For a target moving in a 2D space, the best target localization is expected when $K_{c,t}\geq 3$. For computational cost reduction, at any time $t$ if the target is connected to more than three UWB anchors, $K_{c,t}>3$, we only use measurements with respect to three of the anchors. Let $\mathcal{O}_t$ be the set of measurement $K_{c,t}$ collected at time index $t$ (cardinality of $\mathcal{O}_t$ is $K_{c,t}$). We choose the three anchors to update localization by choosing a measurement vector $\mathbf{o}_t$ with $3$ measurement elements, which takes values from $\mathcal{O}_t$, based on the maximum likelihood value as follows,
\begin{align}
\label{eq::Obs_Prob2}
    \mathbb{P}(\mathbf{o}_t|\mathbf{x}_t) \propto \max_{\mathbf{o}_{t} \in \mathcal{O}_t} \exp(-\frac{1}{2}  (\mathbf{o}_{t} - \Theta(\mathbf{L}\mathbf{x}_t))^{T} \Sigma_{o}^{-1} (\mathbf{o}_{t} - \Theta(\mathbf{L}\mathbf{x}_t)),
\end{align}
where $\Theta(\mathbf{L}\mathbf{x}_t)$ is a vector with elements $\theta_{k}(\mathbf{L}\mathbf{x}_t)$, and $\Sigma_o$ is the diagonal covariance matrix with elements $\sigma_{o}$ according to the measurement vector $\mathbf{o}_t$. It is assumed that measurements are independent and identically distributed (i.i.d).

Because of employing the transition probability and also estimating the MAP most probable trajectory (i.e., the mode sequence from $0$ to $t$), when $K_{c,t}\leq 2$ occasionally, our HMM-based localization algorithm is expected to exhibit some degree of robustness to loss of localization performance. Our experimental demonstrations in Section~\ref{sec::ExpResult} confirm this expectation. On the other hand, Trilateration 
to function needs at least three anchor connections to update the target location estimates~\cite{norrdine2012algebraic}, and cannot operate with $K_c\leq 2$. When $K_{c,t}\leq 2$, depending on the relative location of the target with respect to the anchors', we can have a loss of observability in the extended Kalman filtering framework. This can have a devastating effect on the accuracy of the produced estimates due to the deviation of the system from ideal conditions needed for the validity of linearization models~\cite{chen2003bayesian}.

\medskip
\section{Experimental Demonstrations}
\label{sec::ExpResult}
In this section, we present a set of  experimental evaluations to demonstrate our proposed algorithm's performance.

\subsection{Experimental Setting}
\begin{figure}[!t]
    \centering
    \begin{subfigure}[b]{0.42\textwidth}
        \centering
        \includegraphics[width=\textwidth]{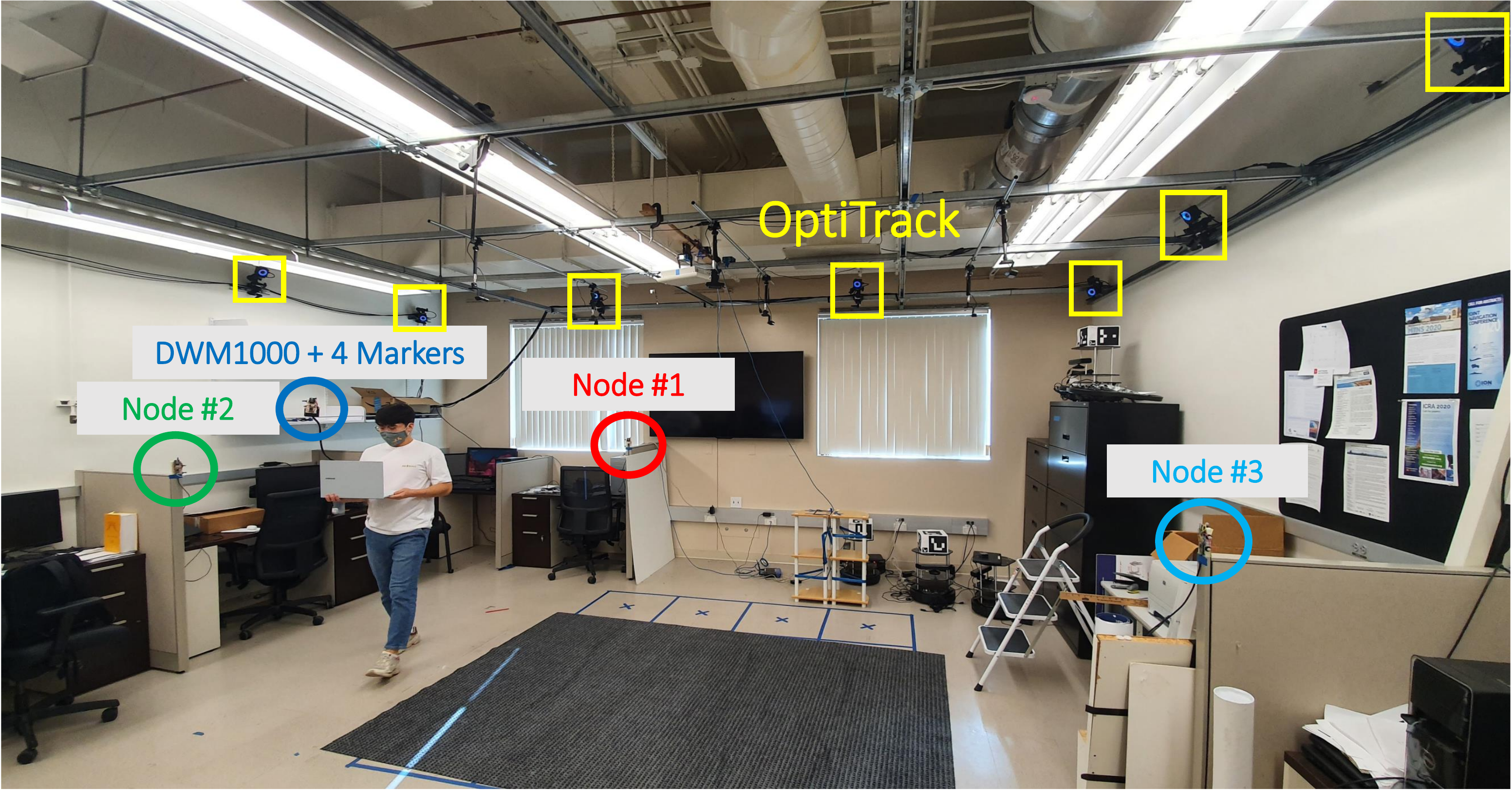}
        {{\small (a) First experiment scenario}}
        \label{fig:3.1}
    \end{subfigure}
    \vskip\baselineskip
    \begin{subfigure}[b]{0.42\textwidth}  
        \centering 
        \includegraphics[width=\textwidth]{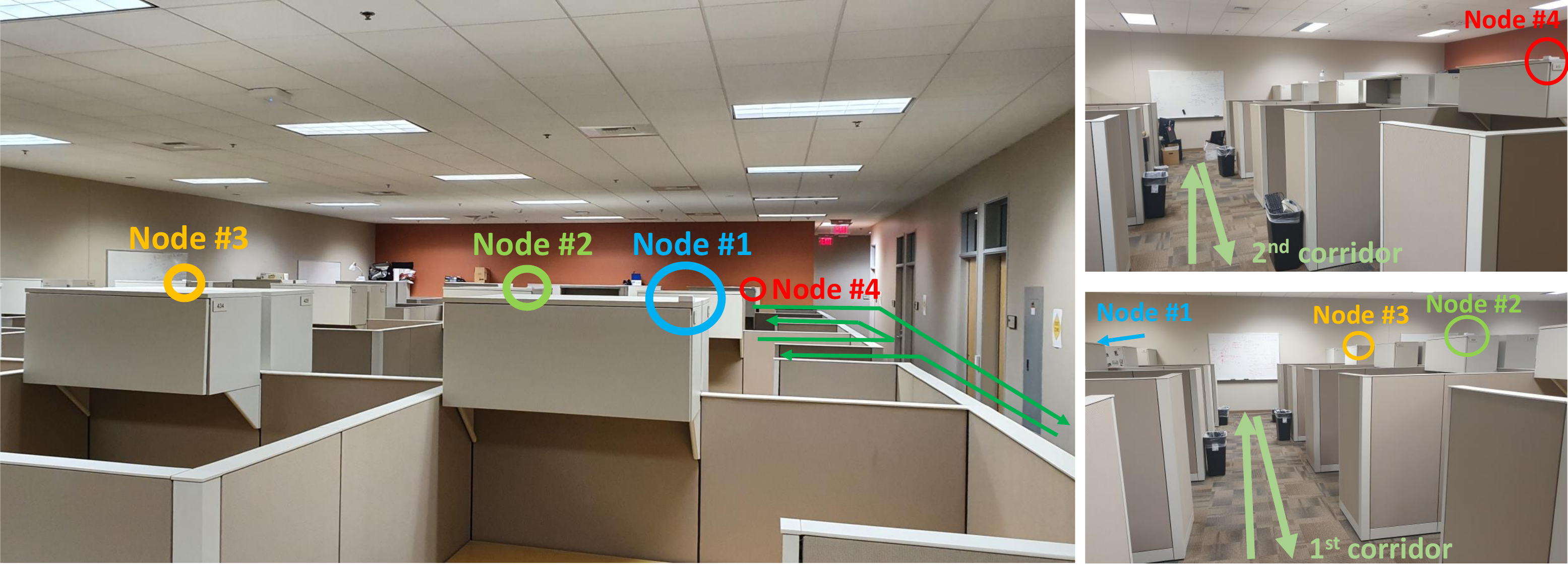}
        {{\small (b) Second experiment scenario}}
    \label{fig:3_2}
    \end{subfigure}
    \caption{{\small The scene of the experiments. The agent walked (a) with DWM1000 in 3 anchors ($8 \: m \times 8 \: m \in \mathbb{R}^2$, $121.5 \: s$), (b) with DWM1001 in 4 anchors ($12 \: m \times 12 \: m \in \mathbb{R}^2$, $59.6 \: s$).}}
    \label{fig:Exp_Environment}
\end{figure}

\begin{figure*}[!t]
    \centering
    \includegraphics[width=0.99\textwidth]{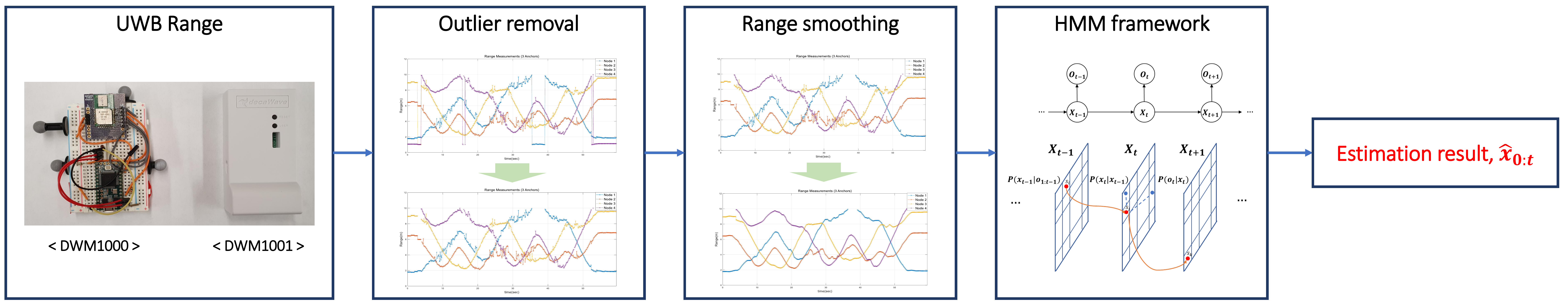}
    \caption{{\small The raw measurements from the UWB sensor are applied to HMM framework after outlier removal and range smoothing. The outliers are removed using a practical threshold on $|o_t-o_{t-1}|$}. Range smoothing is to perform noise smoothing via the moving average procedure using previous range information.}
    \label{fig:Est_Procedure}
\end{figure*}

In our first experiment, a human agent, equipped with a DWM1000 UWB transceiver (tag), walked along an arbitrary trajectory in an indoor environment of size $8\,m\times 8\,m \in \mathbb{R}^2$ with $3$ UWB anchors ($K = 3$) without any obstacle between anchors and the tag (see Fig.~\ref{fig:Exp_Environment}(a)). In practice, LoS measurements can be blocked because of mobile obstacles. In the first experiment, for robustness analysis, at some time instances, we dropped the measurements from node $\#1$ to emulate scenarios that $K_c<3$.
The DWM1000 is paired with a Teensy 3.2 micro-controller for data acquisition. The UWB data communication software is embedded in the micro-controller, and thus the measurement data is logged and synchronized using time stamps. To obtain a high precision reference trajectory data $x_{0:T}^{g}$ for comparison, the optical motion capture system OptiTrack was used. Four OptiTrack markers were placed on the mobile agent for high precision localization. The OptiTrack system consisted of $12$ infrared cameras and was capable of identifying the position of the markers in its measurement zone with an accuracy of $10^{-4} \; m$ and a sampling frequency of $120 \, Hz$. In the second experiment, the DWM1001 UWB transceiver (tag) was placed on the human agent's waist. In this position, it is more likely for
the UWB signals are blocked by the human body resulting in a positive biased range and/or disconnection in some parts of the trajectory. The agent walked along a reference trajectory in an environment of $12\,m\times 12\,m \in \mathbb{R}^2$ with $4$ UWB anchors ($K = 4$) (see Fig.~\ref{fig:Exp_Environment}(b)). For this experiment we did not have access to the OptiTrack system, therefore, we asked the human agent to walk along a geometric reference trajectory. DWM1001 module offers data logging files of range measurement via smartphone Apps (DRTLS Manager). In these experiments, the sampling rate of UWB was set to $10 \, Hz$ ($T_s=0.1 \, sec$). The outliers in the UWB measurement were removed via outlier rejection. Then, the moving average procedure, which provides the unweighted mean of the previous $10$ data points, is employed to perform noise smoothing as shown in Fig.~\ref{fig:Est_Procedure}.

\begin{figure}[!t]
    \centering
    \begin{subfigure}[b]{0.48\textwidth}
        \centering
        \includegraphics[width=\textwidth]{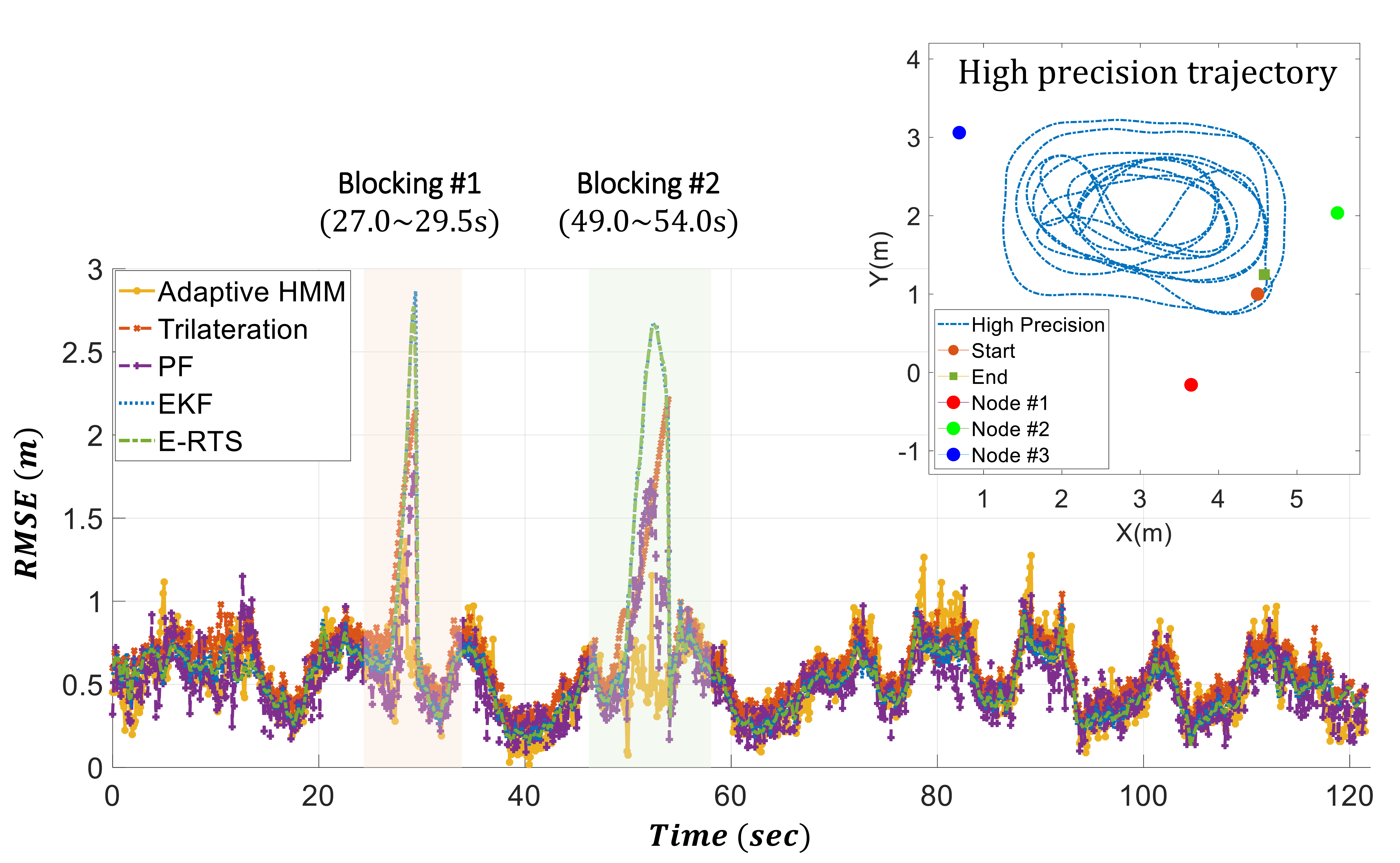}
        {{\small (a) RMSE \& high precision trajectory generated by OptiTrack}}
        \label{fig:Exp_71}
    \end{subfigure}
    \vskip\baselineskip
    \begin{subfigure}[b]{0.48\textwidth}  
        \centering 
        \includegraphics[width=\textwidth]{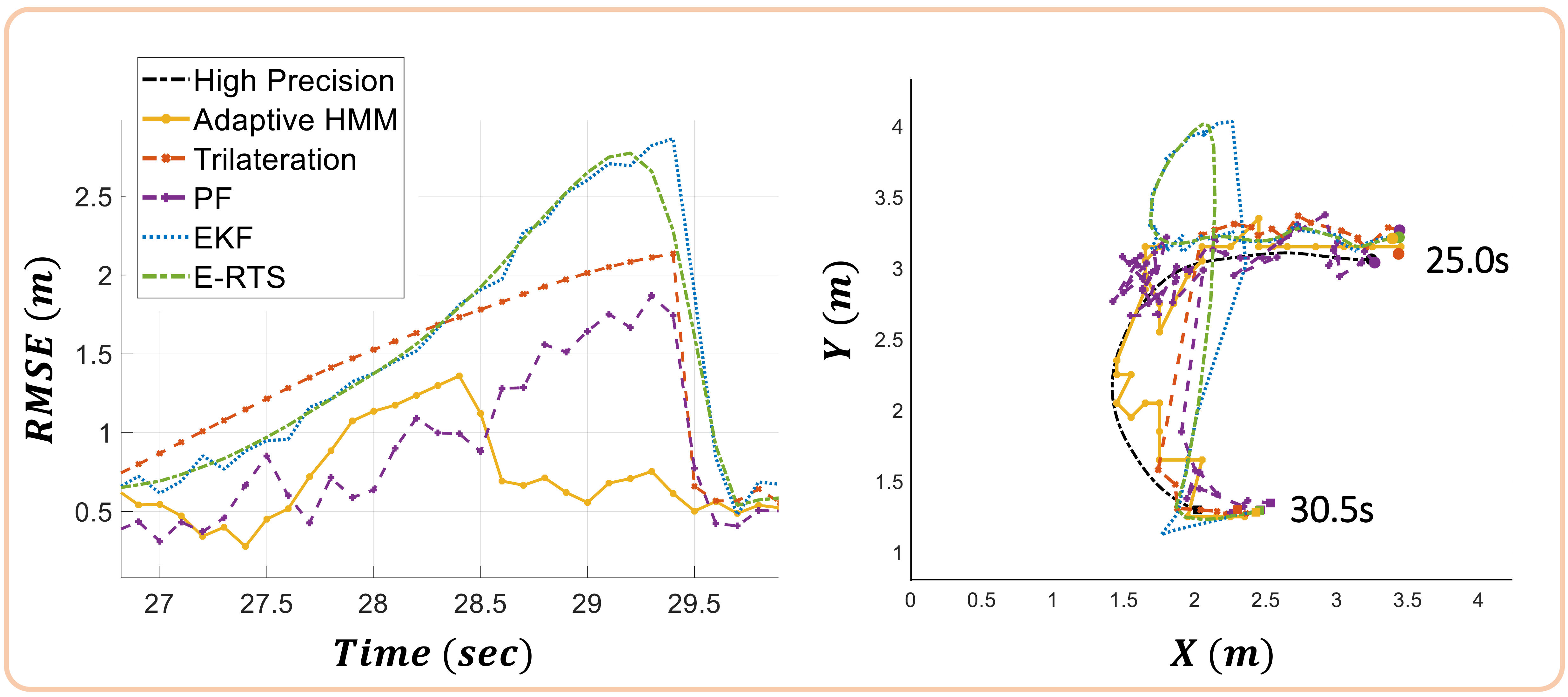}
        {{\small \!\!\!\!(b) RMSE \& estimated trajectories ($o_{1,t}$ dropped in $t\!\in\![27\!,29.5]s$) }}
        \label{fig:Exp_72}
    \end{subfigure}
    \vskip\baselineskip
    \begin{subfigure}[b]{0.48\textwidth}  
        \centering 
        \includegraphics[width=\textwidth]{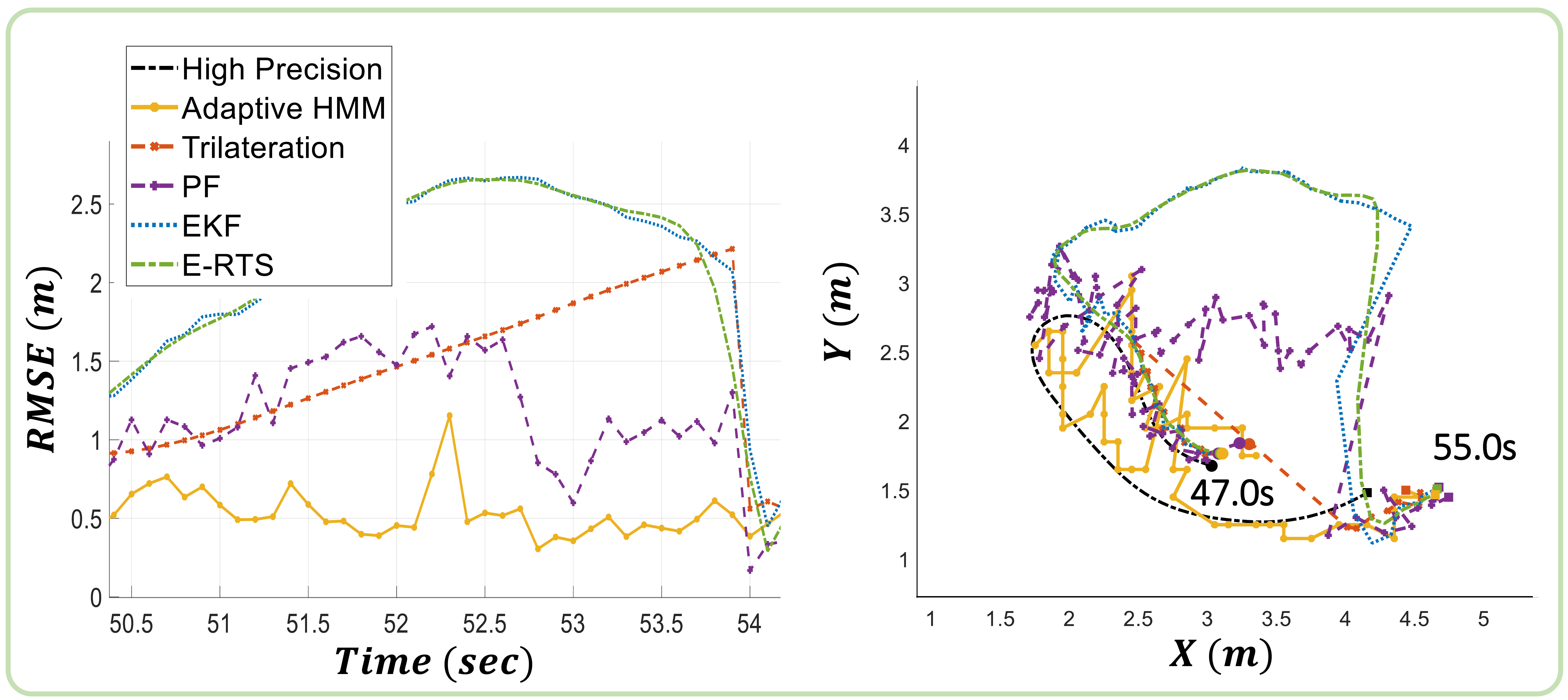}
        {{\small (c) RMSE \& estimated trajectories ($o_{1,t}$ dropped in $t\in[49,54]s$) }}
        \label{fig:Exp_73}
    \end{subfigure}
    \caption{{\small The position RMSE and estimated trajectories according to methods in the first scenario: (a) shows the position RMSE obtained from estimated trajectories and the high precision trajectory from OptiTrack. The UWB signal of Node $\#1$ was dropped ($K_c=2$), thus it makes RMSE be increased during the blocking $\#1$ and $\#2$ periods, (b) shows RMSE and estimated trajectories during blocking $\#1$ from $27.0s$ to $29.5s$, (c) shows RMSE and estimated trajectories during blocking $\#2$ from $49.0s$ to $54.0s$.}}
    \label{fig:Exp_Result2}
\end{figure}

The HMM parameters were set as follows: $\sigma_{x}$ was set to $1.5 \, m$ to consider the movement, the UWB sensor noise covariance was set at $\sigma_{o} = 0.5 \, m$. In resolution tuples, $u_r=u_4=0.1\,m$ was set for high estimation accuracy, and thus other resolution $(u_1,\, u_2,\, u_3)$ are set to $(0.8,\, 0.4,\, 0.2)$ by following the equation~\eqref{eq::Viterbi_eq9}. The size of the experimental places were $8 \: m \times 8 \: m \in \mathbb{R}^2$ and $12 \: m \times 12 \: m \in \mathbb{R}^2$, thus the number of grid was $80 \times 80 = 6,400$, and  $120 \times 120 = 14,400$, respectively.

\subsection{Performance Evaluation}
We compared the performance with Trilateration, which is a recursive least square approach~\cite{norrdine2012algebraic}, Extended Kalman Filter (EKF), Extended Rauch-Tung-Striebel Smoother (E-RTS), and the particle filter (PF)~\cite{doucet2001introduction}. Since we were not using any proprioceptive sensor to enable dead-reckoning, we selected the constant velocity model for target (human) motion, $v_c = 0.5 \; (m/s)$~\cite{scholler2020constant}. For EKF and E-RTS, the initial position was set to $10\;\%$ biased of the true position, and error covariance was set to $0.5\;m$, and $0.5\;m/s$ for $x,y$ position, and velocity, respectively. For PF, the number of particles was set to $N_1$ which is the same as the grid number of the resolution $u_1$. Lastly, we compared computation time, required memory size, and accuracy between the conventional Viterbi algorithm and the proposed one.

\textit{1) First Scenario's Results and Discussion:}
In this scenario, we used the root mean squared error (RMSE) to evaluate the positioning accuracy using high-precision position data. TABLE~\ref{Table_PEA} (1st row) and Fig.~\ref{fig:Exp_Result2} show the position RMSE and estimated trajectories according to different methods, respectively. In Fig.~\ref{fig:Exp_Result2}(b), (c), we observe that the EKF's and E-RTS's estimated trajectories got worse during the time that $K_{c,t}<3$. This is because the Gaussian-based MMSE state estimators result in estimates with large errors when the true posterior becomes multimodal distribution by blocking Node $\#1$ signal ($K_{c,t}=2$). Furthermore, these methods have the disadvantage of requiring initialization (i.e., initial position, error covariance) which can affect the estimation results. PF, which can handle non-Gaussian distribution with finite weighted particles, shows a lower RMSE error than EKF and E-RTS. However, for PF, the MMSE state estimation shows a lower accuracy than the adaptive HMM during the time that $K_{c,t}<3$. In the case of Trilateration, it does not only consider the motion model but also requires 3 connections ($K_c=3$) for estimation. Therefore, an interpolation technique between two points is used for comparison. The adaptive HMM offers MAP trajectory, thus showing the highest robustness against the multimodal posterior.

\begin{figure}[!t]
    \centering
    \includegraphics[width=0.48\textwidth]{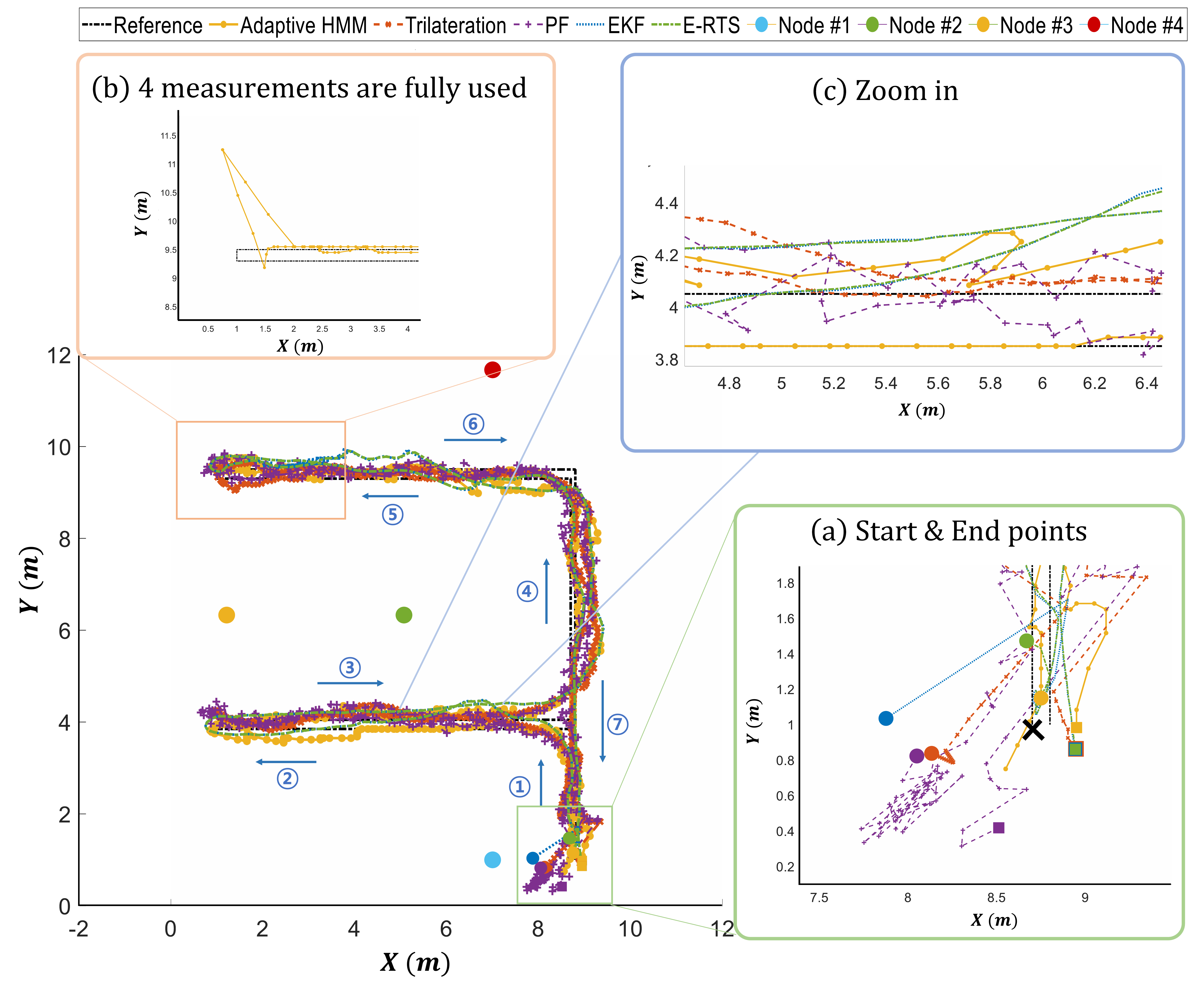}
    \caption{{\small The estimated trajectories are calculated by the adaptive HMM and other methods in the second scenario. The agent walked along the reference trajectory from (1) to (7). (a) shows start and end points. (b) shows estimation error becomes bigger when one of $4$ measurements is a positive biased value. (c) shows estimated trajectories of some sections.}}
    \label{fig:Exp_Result3}
\end{figure}

\textit{2) Second Scenario's Results and Discussion:} 
In this scenario since we did not have the OptiTrack high precision trajectory as reference, we used the Loop Closure Error (LCE) between the start and end point as our performance metric, see Fig.~\ref{fig:Exp_Result3}(a). The reference trajectory is shown in black with the starting point marked as `$\times$'. The experiment was repeated five times to evaluate the average performance. Since the UWB tag is installed in the front part of the agent's waist, obstacles can block UWB connections and/or make the positive biased value. It means that $4$ connections ($K_{c,t}=4$) in some areas, and $3$ connections ($K_{c,t}=3$) in other areas. TABLE~\ref{Table_PEA} (2nd row) shows mean LCE for $5$ experiments according to different methods. We applied the maximum likelihood sets technique~\eqref{eq::Obs_Prob2}, to select $3$ out of $4$ observations to be used in all the methods except Trilateration. The practice of choosing the best $3$ measurements allows us to reduce the effect of employing corrupted measurements due to non-line-of-sight measurements resulting in overall performance improvement, see Fig.~\ref{fig:Exp_Result3}(b). Here, UWB observation data from Node $\#1$ has a positive bias due to obstacles such as desk dividers and the agent's body. In this experiment, we observed the adaptive HMM result is superior to other methods. PF's  estimation result depends on the number of particles; here recall that we use $N_1$ particles in this experiment.

\begin{figure}[!t]
    \centering
    \includegraphics[width=0.48\textwidth]{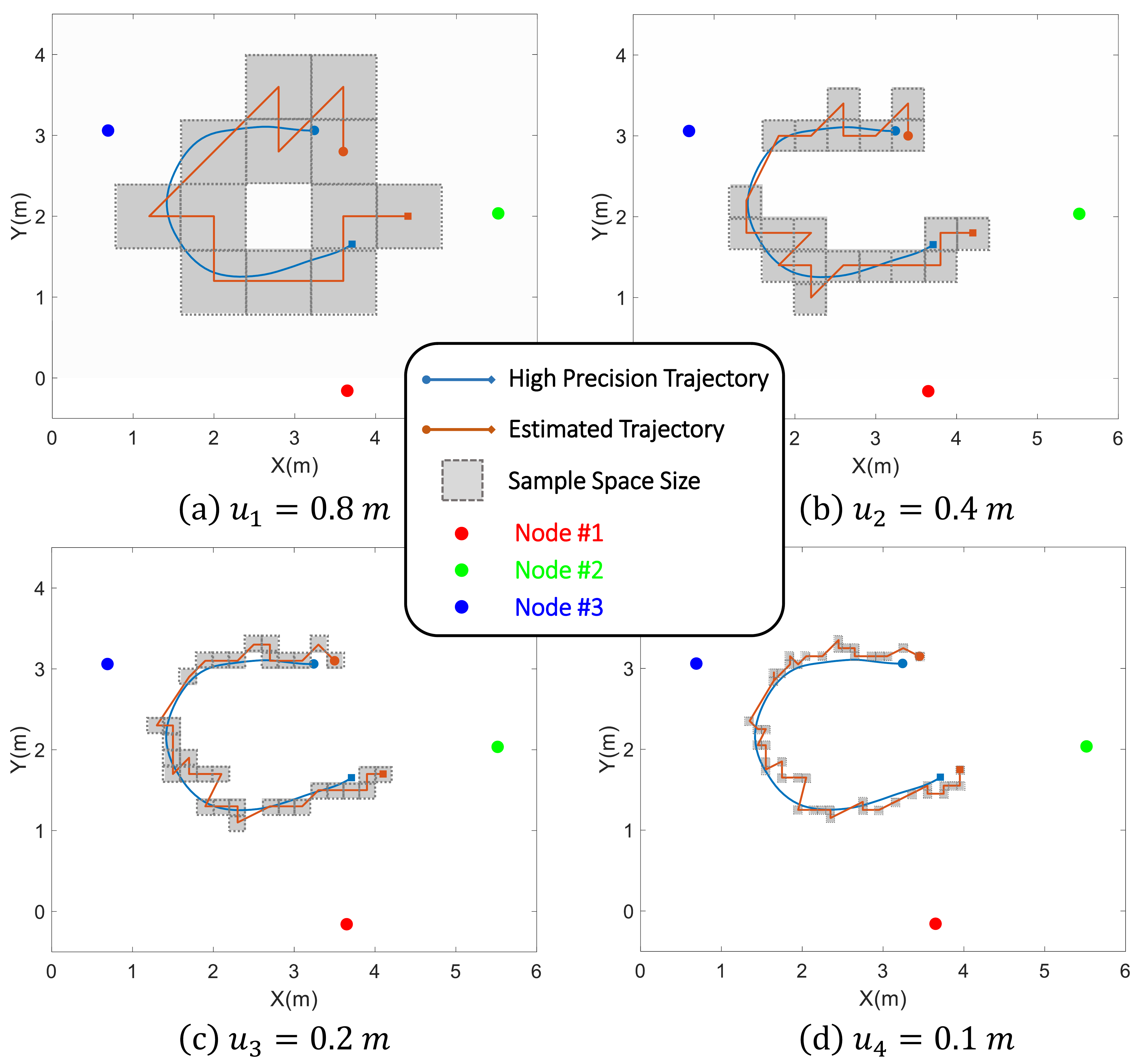}
    \caption{{\small Trajectories are estimated according to the adaptive sample space size, (a) $u_1=0.8\;m$, (b) $u_2=0.4\;m$, (c) $u_3=0.2\;m$, (d) $u_4=0.1\;m$.}}
    \label{fig:Exp_Result1}
\end{figure}

\textit{3) Computation Time, Memory Size, and Accuracy:} We calculated the computation time using a tic-toc function in MATLAB 2022a to compare the conventional Viterbi algorithm and the proposed one. We performed the simulations $10$ times for the proposed one to calculate the mean value of computation time. Also, in order to compare with the required memory size, we simply calculated the number of cells in the memorization table $\Psi_i(\mathbf{x}_{0:T})$. Since the number of cells of $\Psi_i(\mathbf{x}_{0:T})$ is increased critically according to the final time $T$, we only consider the memorization table $\Psi_i(\mathbf{x}_{0:T})$. $\rho_i(\mathbf{x}_t)$ only needs to store values at last time $T$ due to the memoryless property of the dynamic programming. In the first experiment, for the conventional Viterbi, the number of grid is $N=6,400$ based on resolution $0.1\;m$ and total observation data is $1,215$ from (total simulation time, $121.5s$) $\times$ (the sampling rate of UWB, $10Hz$). As a result, required memory size is $6,400 \times 1,215 = 7,776,000$. However, for the proposed one, the number of the lower grid is $N_1=100$ based on resolution $u_1 = 0.8\;m$ and the number of the grid for another tuple is $4$, respectively. As a result, required memory size is $(100 + 4 \times 3) \times 1215 \times 4 = 544,320$. Here, multiplying $4$ means that $4$ trajectories are calculated according to $4$ different sample sizes (see Fig.~\ref{fig:Exp_Result1}). In the same manner, the required memory size for the second scenario are $14,400 \times 596 = 8,582,400$ and $(225 + 4 \times 3) \times 596 \times 4 = 565,008$, respectively. TABLE~\ref{Table_CCTMS} shows that our proposed adaptive method significantly decreased the computation time and memory size of an HMM-based localization. However, as Table~\ref{Table_Compare_other} shows the computational cost can still be higher than the other methods, which in light of the results reported in Table~\ref{Table_PEA} signifies the trade-off between the localization accuracy vs. the computational cost. Finally, we compared accuracy via position RMSE and the difference between two trajectories in the first scenario. The RMSE of the conventional Viterbi is $0.49\;m$, and it is $0.07\;m$ less than the proposed one. The difference between the two trajectories is $0.17\;m$. As a result, we verified the proposed adaptive Viterbi offers an efficient way while maintaining position accuracy.

\begin{table}[h]
\caption{Position estimation accuracy}
\vspace{-0.1in}
\label{Table_PEA}
\begin{center}
\begin{tabular}{||c c c c c c||} 
\hline
& Adaptive HMM & Tri & PF & EKF & E-RTS\\ 
\hline
1st (RMSE, m) & $\mathbf{0.56}$ & 0.64 & 0.57 & 0.61 & 0.60 \\ 
\hline
2nd (LCE, m) & $\mathbf{0.26}$ & 0.81 & 0.51 & 1.08 & 0.67 \\
\hline
\end{tabular}
\end{center}
\end{table}

\begin{table}[h]
\caption{Comparison of computation time/memory size}
\label{Table_CCTMS}
\vspace{-0.1in}
\begin{center}
\begin{tabular}{||c c c||} 
\hline
& Conventional Viterbi & Adaptive one \\ 
\hline
1st & 2433.21 sec / 7,776,000 & $\mathbf{5.95 \: sec / 544,320}$ \\ 
\hline
2nd & $>$ 1 hour / 8,582,400 & $\mathbf{6.97 \: sec/565,008}$  \\
\hline
\end{tabular}
\end{center}
\end{table}

 \begin{table}[h]
 \caption{Comparison of average computation time over $10$ runs}
 \vspace{-0.1in}
 \label{Table_Compare_other}
 \begin{center}
 \begin{tabular}{||c c c c c c||} 
 \hline
 & Adaptive HMM & Tri & PF & EKF & E-RTS\\ 
 \hline
 1st (sec) & 5.81 & 0.15 & 1.35 & 0.03 & 0.05 \\ 
 \hline
 2nd (sec) & 6.81 & 0.09 & 5.69 & 0.05 & 0.07 \\
 \hline
 \end{tabular}
 \end{center}
 \end{table}

\section{Conclusions}
\label{sec::Con}
We proposed an adaptive sample space-based Viterbi algorithm for HMM-based localization in UWB sensor networks. This HMM framework does not impose any restriction on the transition probability or observation probability models and does not also need linearization if the models are nonlinear. To implement this HMM-based localization in an online manner, we applied quad-tree to belief propagation in the Viterbi algorithm. We demonstrated our results using a target localization via range measurements from a UWB anchor network. Our experiments showed that computation time and required memory size decreased significantly.
Future work will focus on tackling biased information in UWB range measurements under NLoS conditions~\cite{Denis2003NlosUWB}. In addition, we will also extend our results to localize a mobile agent in sensor networks where the position of anchors is not known with absolute certainty.

\bibliographystyle{ieeetr}%
\bibliography{bib/Min_RAL.bib}
\end{document}